\documentstyle[a4]{article}
\title{\vspace*{-17mm}
{\bf {\footnotesize In Proceedings of the 15th International Conference
on Computational Linguistics, Coling 94, Kyoto, Japan, pages 748-754.}} \\[3mm]
{\large INCREMENTAL INTERPRETATION: \\
APPLICATIONS, THEORY, AND RELATIONSHIP TO DYNAMIC SEMANTICS\thanks{This
research was supported by the UK Science and Engineering
Research Council, Research Grant RR30718.} \\}}
\date{}
\author{{\large David Milward \& Robin Cooper}\\[4mm]
{\normalsize Centre for Cognitive Science, University of Edinburgh}\\
{\small 2, Buccleuch Place, Edinburgh, EH8 9LW, Scotland,
davidm@cogsci.ed.ac.uk}\\[-3mm]}
\twocolumn
\if@twoside
\else
\fi
\columnwidth \textwidth
\newcommand{\ssp}{\setlength{\baselineskip}{9pt}}
\newcounter{examplectr}
\newcounter{subexamplectr}
\newenvironment{subex}%
   { \addtocounter{subexamplectr}{1}
     \begin{list}
       {\alph{subexamplectr}}%
       {\setlength{\topsep}{-\parskip}
        \setlength{\leftmargin}{0.15in}
        \setlength{\labelsep}{0.1in}}
       \item \ssp
   }%
   {\end{list}}
\newcommand{\myref}[2]{
     (\arabic{#1}{#2})}
\newcommand{\mylabel}[1]{
   \newcounter{#1}
   \setcounter{#1}{\value{examplectr}}}
\newenvironment{ex}%
   { \addtocounter{examplectr}{1}
     \setcounter{subexamplectr}{0}
     \begin{list}
       {\arabic{examplectr})}%
       {\setlength{\topsep}{0.1in}
        \setlength{\leftmargin}{0.25in}
        \setlength{\labelsep}{0.1in}}
       \item
    }%
    {\end{list}}
\begin{document}
\maketitle
\subsection*{\small {\bf ABSTRACT}}

Why should computers interpret language incrementally?  In recent years
psycholinguistic evidence for incremental interpretation has become more and
more compelling, suggesting that humans perform semantic interpretation before
constituent boundaries, possibly word by word. However, possible computational
applications have received less attention.  In this paper we consider various
potential applications, in particular graphical interaction and dialogue. We
then review the theoretical and computational tools available for mapping from
fragments of sentences to fully scoped semantic representations. Finally, we
tease apart the relationship between dynamic semantics and incremental
interpretation.

\subsection*{\small {\bf APPLICATIONS}}

Following the work of, for example, Marslen-Wilson (1973), Just and Carpenter
(1980) and Altmann and Steedman (1988), it has become widely accepted that
semantic interpretation in human sentence processing can occur before sentence
boundaries and even before clausal boundaries.  It is less widely accepted
that there is a need for incremental interpretation in computational
applications.

In the 1970s and early 1980s several computational implementations motivated
the use of incremental interpretation as a way of dealing with structural and
lexical ambiguity (a survey is given in Haddock 1989).  A sentence such as the
following has 4862 different syntactic parses due solely to attachment
ambiguity (Stabler 1991).
\begin{ex}
I put the bouquet of flowers that you gave me for Mothers' Day in the vase
that you gave me for my birthday on the chest of drawers that you gave me
for Armistice Day.
\end{ex}
Although some of the parses can be ruled out using structural preferences
during parsing (such as Late Closure or Minimal Attachment (Frazier 1979)),
extraction of the correct set of plausible readings requires use of real world
knowledge.  Incremental interpretation allows on-line semantic filtering,
i.e.\ parses of initial fragments which have an implausible or anomalous
interpretation are rejected, thereby preventing ambiguities from multiplying
as the parse proceeds.

However, on-line semantic filtering for sentence processing does have
drawbacks. Firstly, for sentence processing using a serial architecture
(rather than one in which syntactic and semantic processing is performed in
parallel), the savings in computation obtained from on-line filtering have to
be balanced against the additional costs of performing semantic computations
for parses of fragments which would eventually be ruled out anyway from purely
syntactic considerations.  Moreover, there are now relatively sophisticated
ways of packing ambiguities during parsing (e.g.\ by the use of
graph-structured stacks and packed parse forests (Tomita 1985)). Secondly, the
task of judging plausibility or anomaly according to context and real world
knowledge is a difficult problem, except in some very limited domains.  In
contrast, statistical techniques using lexeme co-occurrence provide a
relatively simple mechanism which can imitate semantic filtering in many
cases. For example, instead of judging {\it bank} as a financial institution
as more plausible than {\it bank} as a riverbank in the noun phrase {\it the
  rich bank}, we can compare the number of co-occurrences of the lexemes {\bf
  rich} and {\bf bank$_{1}$} (= riverbank) versus {\bf rich} and {\bf
  bank$_{2}$} (= financial institution) in a semantically analysed corpus.
Cases where statistical techniques seem less appropriate are where
plausibility is affected by local context. For example, consider the ambiguous
sentence, {\it The decorators painted a wall with cracks} in the two contexts
{\it The room was supposed to look run-down} vs.\ {\it The clients couldn't
  afford wallpaper}.  Such cases involve reasoning with an interpretation in
its immediate context, as opposed to purely judging the likelihood of a
particular linguistic expression in a given application domain (see e.g.\
Cooper 1993 for discussion).

Although the usefulness of on-line semantic filtering during the processing of
complete sentences is debatable, filtering has a more plausible role to play
in interactive, real-time environments, such as interactive spell checkers
(see e.g.\ Wir\'{e}n (1990) for arguments for incremental parsing in such
environments). Here the choice is between whether or not to have semantic
filtering at all, rather than whether to do it on-line, or at the end of the
sentence.

The concentration in early literature on using incremental interpretation for
semantic filtering has perhaps distracted from some other applications which
provide less controversial applications. We will consider two in detail here:
graphical interfaces, and dialogue.

The Foundations for Intelligent Graphics Project (FIG)\footnote{Joint Councils
  Initiative in Cognitive Science/HCI, Grant 8826213, EdCAAD and Centre for
  Cognitive Science, University of Edinburgh.} considered various ways in
which natural language input could be used within computer aided design
systems (the particular application studied was computer aided kitchen design,
where users would not necessarily be professional designers). Incremental
interpretation was considered to be useful in enabling immediate visual
feedback.  Visual feedback could be used to provide confirmation (for example,
by highlighting an object referred to by a successful definite description),
or it could be used to give the user an improved chance of achieving
successful reference. For example, if sets of possible referents for a
definite noun phrase are highlighted during word by word processing then the
user knows how much or how little information is required for successful
reference.\footnote{This example was inspired by the work of Haddock (1987) on
  incremental interpretation of definite noun phrases. Haddock used an
  incremental constraint based approach following Mellish (1985) to provide an
  explanation of why it is possible to use the noun phrase {\it the rabbit in
    the hat} even when there are two hats, but only one hat with a rabbit in
  it.}

Human dialogue, in particular, task oriented dialogue is characterised by a
large numbers of self-repairs (Levelt 1983, Carletta et al.\ 1993), such as
hesitations, insertions, and replacements. It is also common to find
interruptions requesting extra clarification, or disagreements before the end
of a sentence. It is even possible for sentences started by one dialogue
participant to be finished by another.  Applications involving the
understanding of dialogues include information extraction from conversational
databases, or computer monitoring of conversations.  It also may be useful to
include some features of human dialogue in man-machine dialogue. For example,
interruptions can be used for early signalling of errors and ambiguities.

Let us first consider some examples of self-repair. Insertions add extra
information, usually modifiers e.g.
\begin{ex}
  We start in the middle with ..., in the middle of the paper with a blue disc
  (Levelt 1983:ex.3)
\end{ex}
Replacements correct pieces of information e.g.
\begin{ex}
Go from left again to uh ..., from pink again to blue (Levelt 1983:ex.2)
\end{ex}
In some cases information from the corrected material is incorporated into the
final message.  For example, consider\footnote{Example (a) is reconstructed
  from an actual utterance. Examples (b) and (c) were constructed.} :
\begin{ex}
\begin{subex}
  The three main sources of data come, uh ..., they can be found in the
  references
\end{subex}
\begin{subex}
  John noticed that the old man and his wife, uh ..., that the man got into
  the car and the wife was with him when they left the house
\end{subex}
\begin{subex}
Every boy took, uh ..., he should have taken a water bottle with him
\end{subex}
\end{ex}
\mylabel{semsr} In (a), the corrected material {\it the three main sources of
  data come} provides the antecedent for the pronoun {\it they}.  In (b) the
corrected material tells us that the man is both old and has a wife. In (c),
the pronoun {\it he} is bound by the quantifier {\it every boy}.

For a system to understand dialogues involving self-repairs such as those in
\myref{semsr}{} would seem to require either an ability to interpret
incrementally, or the use of a grammar which includes self repair as a
syntactic construction akin to non-constituent coordination (the relationship
between coordination and self-correction is noted by Levelt (1983)).  For a
system to generate self repairs might also require incremental interpretation,
assuming a process where the system performs on-line monitoring of its output
(akin to Levelt's model of the human self-repair mechanism).  It has been
suggested that generation of self repairs is useful in cases where there are
severe time constraints, or where there is rapidly changing background
information (Carletta, p.c.).

A more compelling argument for incremental interpretation is provided by
considering dialogues involving interruptions.
Consider the following dialogue from the TRAINS corpus (Gross et al., 1993):
\begin{ex}
{\bf A}: so we should move the engine at Avon, \\ engine E, to ...  \\
{\bf B}: engine E1 \\
{\bf A}: E1 \\
{\bf B}: okay \\
{\bf A}: engine E1, to Bath ...
\end{ex}
This requires interpretation by speaker B before the end of A's sentence
to allow objection to the apposition, {\it the engine
at Avon, engine E}. An example of the potential use of interruptions in
human computer interaction is the following:
\begin{ex}
{\bf User}: Put the punch onto ...\\
{\bf Computer}: The punch can't be moved. It's bolted to the floor.
\end{ex}
In this example, interpretation must not only be before the end of the
sentence, but before a constituent boundary (the verb phrase in the user's
command has not yet been completed).

\subsection*{\small {\bf CURRENT TOOLS}}

\subsubsection*{1. Syntax to Semantic Representation}

In this section we shall briefly review work on providing semantic
representations (e.g.\ lambda expressions) word by word.  Traditional layered
models of sentence processing first build a full syntax tree for a sentence,
and then extract a semantic representation from this.  To adapt this to an
incremental perspective, we need to be able to provide syntactic structures
(of some sort) for fragments of sentences, and be able to extract semantic
representations from these.

One possibility, which has been explored mainly within the Categorial Grammar
tradition (e.g. Steedman 1988) is to provide a grammar which can treat most if
not all initial fragments as constituents. They then have full syntax trees
from which the semantics can be calculated.

However, an alternative possibility is to directly link the partial syntax
trees which can be formed for non-constituents with functional semantic
representations. For example, a fragment missing a noun phrase such as {\it
  John likes} can be associated with a semantics which is a function from
entities to truth values. Hence, the partial syntax tree given in
Fig.~1\footnote{The downarrow notation for missing constituents is adopted
  from Synchronous Tree Adjoining Grammar (Shieber \& Schabes 1990).},\\
\begin{verbatim}
       s
      / \
    np  vp
  John  / \
       v   np
    likes
\end{verbatim}
\vspace*{-8mm} \hspace*{23mm} $\downarrow$ \\[3mm]
\hspace*{1cm} {\small{Fig.~1}}
\\[2mm]
can be associated with a semantic representation, \\
{\bf $\lambda$x. likes(john,x)}.

Both Categorial approaches to incremental interpretation and approaches which
use partial syntax trees get into difficulty in cases of left recursion.
Consider the sentence fragment, {\it Mary thinks John}.  A possible partial
syntax tree is provided by Fig.~2.\\
\begin{verbatim}
       s
      / \
    np  vp
  Mary  / \
       v   s
  thinks  / \
         np  vp
        John
\end{verbatim}
\vspace*{-8mm} \hspace*{27mm} $\downarrow$ \\[3mm]
\hspace*{1cm} {\small{Fig.~2}}
\\[2mm]
However, this is not the only possible partial tree. In fact
there are infinitely many different trees possible. The completed
sentence may have an arbitrarily large number of intermediate nodes
between the lower {\bf s} node and the lower {\bf np}.  For example,
{\bf John} could be embedded within a gerund e.g.\  {\it Mary thinks
John leaving here was a mistake}, and this in turn could be embedded
e.g.\  {\it Mary thinks John leaving here being a mistake is
surprising}.  {\bf John} could also be embedded within a sentence
which has a sentence modifier requiring its own {\bf s} node e.g. {\it
Mary thinks John will go home probably}\footnote{The treatment of {\bf
probably} as a modifier of a sentence is perhaps controversial.
However, treatment of it as a verb phrase modifier would merely shift
the potential left recursion to the verb phrase node.}, and this can
be further embedded e.g. {\it Mary thinks John will go home probably
because he is tired}.

The problem of there being an arbitrary number of different partial
trees for a particular fragment is reflected in most current approaches
to incremental interpretation being either incomplete, or not fully
word by word.  For example, incomplete parsers have been proposed by
Stabler (1991) and Moortgat (1988). Stabler's system is a simple
top-down parser which does not deal with left recursive grammars.
Moortgat's M-System is based on the Lambek Calculus: the problem of an
infinite number of possible tree fragments is replaced by a
corresponding problem of initial fragments having an infinite number
of possible types.  A complete incremental parser, which is not fully
word by word, was proposed by
Pulman (1986). This is based on arc-eager left-corner parsing
(see e.g. Resnik 1992).

To enable complete, fully word by word parsing requires a way of
encoding an infinite number of partial trees.  There are several
possibilities. The first is to use a language describing trees where
we can express the fact that {\it John} is dominated by the {\bf s}
node, but do not have to specify what it is immediately dominated by
(e.g. D-Theory, Marcus et al.\ 1983). Semantic representations could
be formed word by word by extracting `default' syntax trees (by
strengthening dominance links into immediated dominance links wherever
possible).

A second possibility is to factor out recursive structures from a
grammar. Thompson et al.\ (1991) show how this can be done for a
phrase structure grammar (creating an equivalent Tree Adjoining
Grammar (Joshi 1987)). The parser for the resulting grammar allows
linear parsing for an (infinitely) parallel system, with the
absorption of each word performed in constant time. At each choice
point, there are only a finite number of possible new partial TAG
trees (the TAG trees represents the possibly infinite number of trees
which can be formed using adjunction).  It should again be possible to
extract `default' semantic values, by taking the semantics from the
TAG tree (i.e.\ by assuming that there are to be no adjunctions).  A
somewhat similar system has recently been proposed by Shieber and
Johnson (1993).

The third possibility is suggested by considering the semantic
representations which are appropriate during a word by word parse.
Although there are any number of different partial trees for the
fragment {\it Mary thinks John}, the semantics of the fragment can be
represented using just two lambda expressions\footnote{Two
representations are appropriate if there are no VP-modifiers as in
dependency grammar.  If VP-modification is allowed, two more
expressions are required:\\ {\bf $\lambda$P. $\lambda$R.
(R($\lambda$x.thinks(x,P(john))))(mary)} and \\{\bf $\lambda$P.
$\lambda$R. $\lambda$Q. Q((R($\lambda$x.thinks(x,P(john))))(mary))}.}:
\begin{quote}
$\lambda$P. thinks(mary,P(john)) \\
$\lambda$P. $\lambda$Q. Q(thinks(mary,P(john)))
\end{quote}
Consider the first. The lambda abstraction (over a functional
item of type {\bf e$\rightarrow$t}) can be thought of as a way of encoding an
infinite set of partial semantic (tree) structures.
For example, the eventual
semantic structure may embed {\bf john} at any depth e.g.
\begin{quote}
thinks(mary,sleeps(john)) \\
thinks(mary,possibly(sleeps(john)))  \\
etc.
\end{quote}
The second expression (a functional item over type
{\bf e$\rightarrow$t} and {\bf t$\rightarrow$t}), allows for eventual
structures where the main sentence is embedded e.g.
\begin{quote}
possibly(thinks(mary,sleeps(john)))
\end{quote}
This third possibility is therefore to provide a syntactic correlate of lambda
expressions.  In practice, however, provided we are only interested in
mapping from a string of words to a semantic representation, and don't
need explicit syntax trees to be constructed, we can merely use the
types of the `syntactic lambda expressions', rather than the
expressions themselves. This is essentially the approach taken in
Milward (1992) in order to provide complete, word by
word, incremental interpretation using
simple lexicalised grammars,
such as a lexicalised version of formal dependency grammar and simple
categorial grammar\footnote{The version of categorial grammar used is
AB Categorial Grammar with Associativity.}.

\subsubsection*{2. Logical Forms to Semantic Filtering}

In processing the sentence {\it Mary introduced
John to Susan}, a word-by-word approach such as Milward (1992)
provides the following logical forms
after the corresponding sentence fragments are
absorbed:
{\small
\begin{tabbing}
Mary introduced John to Suex \= $\lambda x${\bf
intr}({\bf
mary},{\bf john},{\bf sue}) \kill
Mary \> $\lambda$P.P(mary) \\
Mary introduced \> $\lambda$x.$\lambda$y.intr(mary,x,y) \\
Mary introduced John \> $\lambda$y.intr(mary,john,y) \\
Mary introduced John to \> $\lambda$y.intr(mary,john,y) \\
Mary introduced John to Sue\> intr(mary,john,sue)
\end{tabbing}}
\noindent
Each input level representation is appropriate for the meaning of an
incomplete sentence, being either a proposition or a function into a
proposition.

In Chater et al.\ (1994) it is argued that the incrementally derived
meanings are not judged for plausibility directly, but instead are
first turned into existentially quantified propositions. For example,
instead of judging the plausibility of {\bf
$\lambda$x.$\lambda$y.intr(mary,x,y)}, we judge the plausibility of
{\bf $\exists$(x,T,$\exists$(y,T,intr(mary,x,y)))}\footnote{The proposition
{\bf T} is always true. See Chater et al.\ (1994) for
discussion of whether it is more appropriate to use a non-trivial
restrictor.}.  This is just the proposition {\it Mary introduced
something to something} using a generalized quantifier notation
of the form {\bf Quantifier(Variable,Restrictor,Body)}.

Although the lambda expressions are built up monotonically, word by word,
the propositions formed from them may need to be retracted, along with
all the resulting inferences. For example,
{\it Mary introduced something
to something} is inappropriate if the final sentence is
{\it Mary introduced noone to anybody}.
A rough algorithm
is as follows: \\[3mm]
1. Parse a new word, Word$_{i}$ \\[1mm]
2. Form a new lambda expression by combining the lambda expression formed after
parsing Word$_{i-1}$ with the lexical semantics for Word$_{i}$ \\[1mm]
3. Form a proposition, P$_{i}$, by existentially quantifying
over the lambda abstracted variables.\\[1mm]
4. Assert P$_{i}$. If P$_{i}$ does not entail P$_{i-1}$ retract
P$_{i-1}$ and all conclusions made from it\footnote{Retraction can be
performed by using a tagged database, where each proposition
is paired with a set
of sources e.g. given {\bf (P$\rightarrow$Q,\{u4\})}, and {\bf (P,\{u5\})}
then{\bf (Q,\{u4,u5\})} can be deduced.}.\\[1mm]
5. Judge the plausibility of P$_{i}$. If implausible block this derivation.
\\[3mm]
It is worth noting that the need for retraction is not due to a
failure to extract the correct `least commitment' proposition from the
semantic content of the fragment {\it Mary introduced}. This is due to
the fact that it is possible to find pairs of possible continuations
which are the negation of each other (e.g. {\it Mary introduced noone
to anybody} and {\it Mary introduced someone to somebody}).  The only
proposition compatible with both a proposition, {\bf p}, and its
negation, {\bf $\neg$p} is the trivial proposition, {\bf T} (see Chater et
al. for further discussion).

\subsubsection*{3. Incremental Quantifier Scoping}

So far we have only considered semantic representations which do not
involve quantifiers (except for the existential quantifier introduced
by the mechanism above).

In sentences with two or more quantifiers, there is generally an ambiguity
concerning which quantifier has wider scope.  For example, in
sentence (a) below the preferred
reading is for the same kid to have climbed every tree (i.e.\ the
universal quantifier is within the scope of the existential) whereas
in sentence (b) the preferred reading is where the universal
quantifier has scope over the existential.
\begin{ex}
\begin{subex}
A tireless kid climbed every tree.
\end{subex}
\begin{subex}
There was a fish on every plate.
\end{subex}
\label{quant}
\end{ex}
Scope preferences sometimes seem to be established before the end of a
sentence.  For example, in sentence (a) below, there seems a
preference for an outer scope reading for the first quantifier as soon
as we interpret {\it child}. In (b) the preference, by the time we get
to e.g.\ {\it grammar}, is for an inner scope reading for the first
quantifier.
\begin{ex}
\begin{subex}
A teacher gave every child a great deal of homework on
grammar.
\end{subex}
\begin{subex}
Every girl in the class showed a rather strict new teacher
the results
of her attempt to get the grammar exercises correct.
\end{subex}
\label{incquant}
\end{ex}
This intuitive evidence can be backed up by considering garden
path effects
with quantifier scope ambiguities (called
{\it jungle paths} by Barwise 1987). The original examples, such
as the
following,
\begin{ex}
Statistics show that every 11 seconds a man is mugged here in
New York city.
We are here today to interview him
\end{ex}
showed that preferences for a particular scope are established and are
overturned. To show that preferences are sometimes established before
the end of a sentence, and before a potential sentence end, we
need to show garden path effects in examples such as the following:
\begin{ex}
Mary put the information that statistics show that every 11 seconds a man
is mugged here in New York city and that she was to interview him
in her diary
\end{ex}
Most psycholinguistic experimentation has been concerned with which
scope preferences are made, rather than the point at which the preferences
are established (see e.g.\ Kurtzman and MacDonald, 1993).
Given the intuitive evidence, our hypothesis is that scope preferences
can sometimes be established early, before the end of a sentence. This
leaves open the possibility that in other cases, where the scoping information
is not particularly of interest to the hearer, preferences are determined
late, if at all.

\subsubsection*{3.1 Incremental Quantifier Scoping: Implementation}

Dealing with quantifiers incrementally is a rather similar problem to
dealing with fragments of trees incrementally. Just as it is
impossible to predict the level of embedding of a noun phrase such as
{\it John} from the fragment {\it Mary thinks John}, it is also
impossible to predict the scope of a quantifier in a fragment with
respect to the arbitrarily large number of quantifiers which
might appear later in the sentence.
Again the problem can be avoided by a form of packing.
A particularly simple way of doing this is to use unscoped logical
forms where quantifiers are left in situ (similar to the representations
used by Hobbs and Shieber (1987), or to Quasi Logical Form (Alshawi 1990)).
For example, the fragment {\it Every man gives a book} can be given the
following representation:
\begin{ex}
$\lambda$z.gives($<\forall$,x,man(x)$>$,$<\exists$,y,book(y)$>$,z)
\end{ex}
Each quantified term consists of a quantifier, a variable and a restrictor,
but no body.  To convert lambda expressions to unscoped propositions, we
replace an occurrence of each argument with an empty existential quantifier
term. In this case we obtain:
\begin{ex}
gives($<\forall$,x,man(x)$>$,$<\exists$,y,book(y)$>$,$<\exists$,z,T$>$)
\end{ex}
Scoped propositions can then be obtained by using an
outside-in quantifier scoping algorithm (Lewin, 1990), or an
inside-out algorithm with a free variable constraint (Hobbs and
Shieber, 1987).  The propositions formed can
then be judged for plausibility.

To imitate jungle path phenomena, these plausibility judgements need
to feed back into the scoping procedure for the next fragment.  For
example, if {\it every man} is taken to be scoped outside {\it a book}
after processing the fragment {\it Every man gave a book}, then this
preference should be preserved when determining the scope for the full
sentence {\it Every man gave a book to a child}.  Thus instead of
doing all quantifier scoping at the end of the sentence,
each new quantifier is scoped relative to the
existing quantifiers (and operators such as negation, intensional
verbs etc.). A preliminary implementation achieves this by annotating
the semantic representations with node names, and recording which quantifiers
are `discharged' at which nodes, and in which order.

\subsection*{\small {\bf DYNAMIC SEMANTICS}}

Dynamic semantics adopts the view that ``the meaning of a sentence
does not lie in its truth conditions, but rather in the way in which
it changes (the representation of) the information of the interpreter''
(Groenendijk and Stokhof, 1991).  At first glance such a view seems
ideally suited to incremental interpretation.  Indeed, Groenendijk and
Stokhof claim that the compositional nature of Dynamic
Predicate Logic enables one to ``interpret a text in an on-line
manner, i.e., incrementally, processing and interpreting each basic
unit as it comes along, in the context created by the interpretation
of the text so far''.

Putting these two quotes together is, however, misleading, since it
suggests a more direct mapping between incremental semantics and
dynamic semantics than is actually possible.  In an incremental
semantics, we would expect the information state of an interpreter to
be updated word by word.  In contrast, in dynamic semantics, the
order in which states are updated is determined by semantic structure,
not by left-to-right order (see e.g.\ Lewin, 1992 for discussion).
For example, in Dynamic Predicate Logic (Groenendijk \& Stokhof,
1991), states are threaded from the antecedent of a conditional into
the consequent, and from a restrictor of a quantifier into the body.
Thus, in interpreting,
\begin{ex}
John will buy it right away, if a car impresses him
\end{ex}
\mylabel{cat}
the input state for evaluation of {\it John will buy it right away} is
the output state from the antecedent {\it a car impresses him}.
In this case the
threading through semantic structure is in the opposite order to the order in
which the two clauses appear in the sentence.

Some intuitive justification for the direction of threading in dynamic
semantics is provided by considering appropriate orders for evaluation of
propositions against a database: the natural order in which to evaluate a
conditional is first to add the antecedent, and then see if the consequent can
be proven.  It is only at the sentence level in simple narrative texts that
the presentation order and the natural order of evaluation necessarily
coincide.

The ordering of anaphors and their antecedents is often used informally to
justify left-to-right threading or threading through semantic structure.
However, threading from left-to-right disallows examples of optional
cataphora, as in example \myref{cat}{}, and examples of compulsory cataphora
as in:
\begin{ex}
Beside her, every girl could see a large crack
\end{ex}
Similarly, threading from the antecedents of conditionals into the
consequent fails for examples such as:
\begin{ex}
Every boy will be able to see out of a window if he wants to
\end{ex}
It is also possible to get sentences with `donkey' readings,
but where the indefinite is in the consequent:
\begin{ex}
A student will attend the conference if we can get together enough
money for her air fare
\end{ex}
\mylabel{airfare}
This sentence seems to get a reading where we are not talking about a
particular student (an outer existential), or about a typical student
(a generic reading). Moreover, as noted by Zeevat (1990), the use of
any kind of {\it ordered} threading will tend to fail for
Bach-Peters sentences, such as:
\begin{ex}
Every man who loves her appreciates a woman who lives with him
\end{ex}
\mylabel{bach}
For this kind of example, it is still possible to use a standard
dynamic semantics, but only if there is some prior level of reference
resolution which reorders the antecedents and anaphors appropriately.
For example, if \myref{bach}{} is converted into the
`donkey' sentence:
\begin{ex}
Every man who loves a woman who lives with him appreciates her
\end{ex}

When we consider threading of possible worlds, as in Update Semantics
(Veltman 1990), the need to distinguish between the order of
evaluation and the order of presentation becomes more clear cut.
Consider trying to perform threading in left-to-right order during
interpretation of the sentence, {\it John left if Mary left}.  After
processing the proposition {\it John left} the set of worlds is
refined down to those worlds in which John left. Now consider
processing {\it if Mary left}. Here we want to reintroduce some
worlds, those in which neither Mary or John left.  However, this is
not allowed by Update Semantics which is {\it eliminative}: each new
piece of information can only further refine the set of worlds.

It is worth noting that the difficulties in trying to combine
eliminative semantics with left-to-right threading apply to
constraint-based semantics as well as to Update Semantics.  Haddock
(1987) uses incremental refinement of sets of possible referents. For
example, the effect of processing {\it the rabbit} in the noun phrase
{\it the rabbit in the hat} is to provide a set of all rabbits. The
processing of {\it in} refines this set to rabbits which are in
something. Finally, processing of {\it the hat} refines the set to
rabbits which are in a hat.  However, now consider processing {\it the
rabbit in none of the boxes}.  By the time {\it the rabbit in} has
been processed, the only rabbits remaining in consideration are
rabbits which are in something.  This incorrectly rules out the
possibility of the noun phrase referring to a rabbit which is in
nothing at all. The case is actually a parallel to the earlier example
of {\it Mary introduced someone to something} being inappropriate if
the final sentence is {\it Mary introduced noone to anybody}.

Although this discussion has argued that it is not possible to thread
the states which are used by a dynamic or eliminative semantics from
left to right, word by word, this should not be taken as an argument
against the use of such a semantics in incremental interpretation.
What is required is a slightly more indirect approach.  In the present
implementation, semantic structures (akin to logical forms) are built
word by word, and each structure is then evaluated independently using
a dynamic semantics (with threading performed according to the
structure of the logical form).

\subsection*{\small {\bf IMPLEMENTATION}}

At present there is a limited implementation, which performs a mapping
from sentence fragments to fully scoped logical representations. To
illustrate its operation, consider the following discourse:
\begin{ex}
London has a tower. Every parent shows it ...
\end{ex}
We assume that the first sentence has been processed, and concentrate on
processing the fragment. The implementation consists of five modules: \\[2mm]
1. A word-by-word incremental parser for a lexicalised version of dependency
grammar (Milward, 1992). This takes fragments of sentences
and maps them to unscoped logical forms. \\[1mm]
INPUT: {\bf Every parent shows it} \\
OUTPUT: {\small
{\bf $\lambda$z.show($<\forall$,x,parent(x)$>$,$<$pronoun,y$>$,z)}} \\[2mm]
2. A module which replaces lambda abstracted variables with existential
quantifiers in situ. \\[1mm]
INPUT: Output from 1. \\
OUTPUT: {\small
{\bf show($<\forall$,x,parent(x)$>$,$<$pronoun,y$>$,$<\exists$,z,T$>$)}}
\\[2mm]
3. A pronoun coindexing procedure which replaces pronoun variables with
a variable from the same sentence, or from the preceding context. \\[1mm]
INPUT: Output(s) from 2 and a list of variables available from the context. \\
OUTPUT: {\small {\bf show($<\forall$,x,parent(x)$>$,w,$<\exists$,z,T$>$)}}
\\[2mm]
4. An outside-in quantifier scoping algorithm based on Lewin (1990).\\[1mm]
INPUT: Output from 3. \\
{\small OUTPUT1}:
{\small {\bf $\forall$(x,parent(x),$\exists$(z,T,show(x,w,z)))}} \\
{\small OUTPUT2}:
{\small {\bf $\exists$(z,T,$\forall$(x,parent(x),show(x,w,z)))}}
\\[2mm]
5. An `evaluation' procedure based on Lewin (1992), which takes a logical
form containing free variables
(such as the {\it w} in the LF above), and evaluates it using a
dynamic semantics in the context given by the preceding sentences. The output
is a new logical form representing the context as a whole, with all
variables correctly bound. \\[1mm]
INPUT: Output(s) from 4, and the context, \\
{\bf $\exists$(w,T,tower(w) \& has(london,w))} \\
{\small OUTPUT1}: {\small {\bf $\exists$(w,T,tower(w) \& has(london,w) \\ \&
$\forall$(x,parent(x),$\exists$(z,T,show(x,w,z))))}}
\\
{\small OUTPUT2}: {\small {\bf $\exists$(w,T,$\exists$(z,T,tower(w)
\& has(london,w)
\& $\forall$(x,parent(x),show(x,w,z))))}}
\\[2mm]
At present, the coverage of module 5 is limited, and module 3 is
a naive coindexing procedure which allows a pronoun to be coindexed with any
quantified variable or proper noun in the context or the current sentence.

\subsection*{\small {\bf CONCLUSIONS}}

The paper described some potential applications of incremental
interpretation.  It then described the series of steps required in
mapping from initial fragments of sentences to propositions which can
be judged for plausibility. Finally, it argued that the apparently
close relationship between the states used in incremental semantics
and dynamic semantics fails to hold below the sentence level, and
briefly presented a more indirect way of using dynamic semantics in
incremental interpretation.

\subsection*{\small {\bf REFERENCES}}

\noindent
 Alshawi, H. (1990). Resolving Quasi Logical Forms. {\it Computational
Linguistics, 16}, p.133-144.
\\[2mm]
Altmann, G.T.M. and M.J.~Steedman (1988). Interaction with Context during
Human Speech Comprehension. {\it Cognition, 30}, p.191-238.
\\[2mm]
Barwise, J. (1987). Noun Phrases, Generalized Quantifiers and Anaphors.
In P. Gardenfors, Ed., {\it Generalized Quantifiers}, p.1-29,
Dordrecht: Reidel.
\\[2mm]
Carletta, J., R.~Caley and S.~Isard (1993). A Collection of Self-repairs
from the Map Task Corpus. Research Report, HCRC/TR-47, University of Edinburgh.
\\[2mm]
Chater, N., M.J.~Pickering and D.R.~Milward (1994). What is Incremental
Interpretation? ms. To appear in Edinburgh Working Papers in Cognitive
Science.
\\[2mm]
Cooper, R. (1993). A Note on the Relationship between Linguistic
Theory and Linguistic Engineering. Research Report, HCRC/RP-42,
University of Edinburgh.
\\[2mm]
Frazier, L. (1979).  {\it On Comprehending Sentences: Syntactic
Parsing Strategies}. Ph.D. Thesis, University of Connecticut.
Published by Indiana University
Linguistics Club.
\\[2mm]
Groenendijk, J. and M.~Stokhof (1991). Dynamic Predicate Logic.
{\it Linguistics and Philosophy, 14}, p.39-100.
\\[2mm]
Gross, D., J.~Allen and D.~Traum (1993). The TRAINS 91 Dialogues.
TRAINS Technical Note 92-1, Computer Science Dept., University of Rochester.
\\[2mm]
Haddock, N.J. (1987). Incremental semantic interpretation and
incremental syntactic analysis.  Ph.D.~Thesis, University of Edinburgh.
\\[2mm]
Haddock, N.J. (1989). Computational Models of Incremental Semantic
Interpretation. {\it Language and Cognitive Processes, 4, (3/4)},
Special Issue, p.337-368.
\\[2mm]
Hobbs, J.R. and S.M.~Shieber (1987). An Algorithm for Generating
Quantifier Scoping. {\it Computational Linguistics, 3}, p47-63.
\\[2mm]
Joshi, A.K. (1987). An Introduction to Tree Adjoining Grammars. In
Manaster-Ramer, Ed., {\it Mathematics of Language}, Amsterdam:
John Benjamins.\\[2mm]
Just, M. and P.~Carpenter (1980). A Theory of Reading from Eye Fixations
to Comprehension. {\it Psychological Review, 87}, p.329-354.
\\[2mm]
Kurtzman, H.S. and M.C.~MacDonald (1993). Resolution of Quantifier Scope
Ambiguities. {\it Cognition, 48(3)}, p.243-279.
\\[2mm]
Lewin, I. (1990). A Quantifier Scoping Algorithm without a Free Variable
Constraint.
In {\it Proceedings of COLING 90}, Helsinki, vol 3, p.190-194.
\\[2mm]
Lewin, I. (1992). Dynamic Quantification in Logic and Computational Semantics.
Research report, Centre for Cognitive Science, University of Edinburgh.
\\[2mm]
Levelt, W.J.M. (1983). Modelling and Self-Repair in Speech.
{\it Cognition, 14}, p.41-104.
\\[2mm]
Marcus, M., D.~Hindle, and M.~Fleck (1983). D-Theory: Talking about
Talking about Trees. In {\it Proceedings of the 21st ACL}, Cambridge, Mass.
p.129-136.
\\[2mm]
Marslen-Wilson, W. (1973). Linguistic Structure and Speech Shadowing at
Very Short Latencies. {\it Nature, 244}, p.522-523.
\\[2mm]
Mellish, C.S. (1985). {\it Computer Interpretation of Natural Language
Descriptions}. Chichester: Ellis Horwood.
\\[2mm]
Milward, D.R. (1991). Axiomatic Grammar, \\[2mm] Non-Constituent
Coordination, and
Incremental Interpretation. Ph.D.~Thesis, University of Cambridge.
\\[2mm]
Milward, D.R. (1992). Dynamics, Dependency Grammar and Incremental
Interpretation.
In {\it Proceedings of COLING 92}, Nantes, vol 4, p.1095-1099.
\\[2mm]
Moortgat, M. (1988). {\it Categorial Investigations: Logical and Linguistic
Aspects of the Lambek Calculus}, Dordrecht: Foris.
\\[2mm]
Pulman, S.G. (1986). Grammars, Parsers, and Memory Limitations. {\it
Language
and Cognitive Processes, 1(3)}, p.197-225.
\\[2mm]
Resnik, P. (1992). Left-corner Parsing and Psychological Plausibility.
In {\it Proceedings of COLING 92}, Nantes, vol 1, p.191-197.
\\[2mm]
Shieber, S.M. and M.~Johnson (1993). Variations on Incremental Interpretation.
{\it Journal of Psycholinguistic Research, 22(2)}, p.287-318.
\\[2mm]
Shieber, S.M. and Y.~Schabes (1990). Synchronous Tree-Adjoining Grammars.
In {\it Proceedings of COLING 90}, Helsinki, vol 3, p.253-258.
\\[2mm]
Stabler, E.P. (1991).  Avoid the pedestrian's paradox.  In R.~Berwick,
S.~Abney, and C.~Tenny, Eds., {\it Principle-Based Parsing:
Computation and Psycholinguistics}. Kluwer.
\\[2mm]
Steedman, M. (1988). Combinators and Grammars. In R.~Oehrle et al., Eds.,
{\it Categorial Grammars and Natural Language Structures}, p.417-442.
\\[2mm]
Thompson, H., M.~Dixon, and J.~Lamping (1991). Compose-Reduce Parsing.
In {\it Proceedings of the 29th ACL}, p.87-97.
\\[2mm]
Tomita, M. (1985). {\it Efficient Parsing for Natural Language}. Kluwer.
\\[2mm]
Veltman F. (1990). Defaults in Update Semantics. In H.~Kamp, Ed.,
{\it Conditionals, Defaults and Belief Revision}, DYANA Report 2.5.A,
Centre for Cognitive Science, University of Edinburgh.
\\[2mm]
Wir\'{e}n, M. (1990). Incremental Parsing and Reason Maintenance. In
{\it Proceedings of COLING 90}, Helsinki, vol 3, p.287-292.
\\[2mm]
Zeevat, H. (1990). Static Semantics. In J.~van Benthem, Ed.,
{\it Partial and Dynamic Semantics~I}, DYANA Report 2.1.A,
Centre for Cognitive Science, University of Edinburgh.
\end{document}